\documentclass{article}
\usepackage{spconf,amsmath,epsfig,amssymb,amsthm}
\usepackage{graphicx}
\usepackage{cite}
\usepackage{multirow}
\usepackage{subfigure}
\graphicspath{{./Figures/}}
\DeclareGraphicsExtensions{.pdf,.jpg,.png,.gif}
\usepackage{paralist, tabularx}

\title{Buried object detection from B-scan ground penetrating radar data using Faster-RCNN}

\name {Minh-Tan Pham, S\'ebastien Lef\`evre \thanks{This work is supported by the R\'egion Bretagne grant.}}
\address{
Universit\'e Bretagne Sud - IRISA, UMR 6074, F-56000 Vannes - France\\
\texttt{\{minh-tan.pham,sebastien.lefevre\}@irisa.fr}}

\begin{document}
%
\maketitle
\begin{abstract}
In this paper, we adapt the Faster-RCNN framework for the detection of underground buried objects (i.e. hyperbola reflections) in B-scan ground penetrating radar (GPR) images. Due to the lack of real data for training, we propose to incorporate more simulated radargrams generated from different configurations using the gprMax toolbox. Our designed CNN is first pre-trained on the grayscale Cifar-10 database. Then, the Faster-RCNN framework based on the pre-trained CNN is trained and fine-tuned on both real and simulated GPR data. Preliminary detection results show that the proposed technique can provide significant improvements compared to classical computer vision methods and hence becomes quite promising to deal with this kind of specific GPR data even with few training samples.
\end{abstract}
%

\section{Introduction}
\label{sec:intro}
Ground penetrating radar (GPR) is one of the most widely used geophysical techniques applied to detect underground buried objects such as landmines, pipes or archaeological artifacts, etc. A GPR system transmits an electromagnetic wave into the ground at several spatial positions and receives the reflected signal to form the subsurface 2-D high resolution image (a B-scan radargram). Within such images, underground objects appear particularly as hyperbolic-shaped signatures. The detection of buried objects can be therefore considered as the detection of reflected hyperbolas in GPR images, which has been tackled using image recognition and computer vision techniques so far in the literature.

Different unsupervised and supervised approaches have been investigated to perform automatic detection of buried objects using GPR B-scan images. One of the most popular and classical approaches is the Hough transform (HT) method. In \cite{windsor2014data,li2016tree}, the generalized HT and the randomized HT were employed to find the hyperbola parameters which are recorded within the Hough accumulator space. However, most of these Hough-based approaches are still limited to the fact that the handling and discretization of a great number of parameters could lead to a huge computational complexity. In \cite{sagnard2016template, terrasse2016automatic}, the template matching and the dictionary-based techniques were used to determine hyperbola signatures. These methods are based on the correlation score between each GPR image patch and the template or the dictionary model. They again require many parameters for the setup and definition of different templates or dictionary models. Some other methods were proposed using supervised pattern recognition approach such as the HOG feature-based classification \cite{torrione2014histograms}, the Viola-Jones learning algorithm based on Haar-like features \cite{maas2013using}. However, their results still involve several unexpected false alarms and missed detection targets. Thus, the detection of hyperbolas from GPR images using these classical recognition strategies is still considered as a challenging task.

Recently, the incredible development of deep convolutional neural networks (CNNs) in computer vision domain has provided a lot of tools and frameworks to tackle various tasks in image understanding and recognition. Efforts have been also done in GPR image processing. In \cite{besaw2015deep}, deep CNNs were exploited for classifying B-scan profiles into threat and non-threat classes.  In \cite{lameri2017landmine}, landmine detection from GPR data using CNNs has provided quite promising results compared to other computer vision methods. In \cite{reichman2017some}, the authors discussed some good practices when applying CNNs to the detection of burried threats from GPR data. However, all of the deep learning based techniques in the literature have focused on classification or patch-based detection using sliding window over the whole image. In this work, we would like to perform an end-to-end framework for hyperbola detection from GPR B-scan. To do this, we apply the Faster-RCNN framework \cite{ren2015faster} which has proved very effective performance in the computer vision domain. The contributions of the paper are threefolds: 1) we first define and pre-train a CNN using the grayscale Cifar-10 database and then transfer the weights into the Faster-RCNN framework; 2) the training data are created partly from the real collected GPR acquisitions and partly from the simulated radargrams obtained by gprMax toolbox \cite{warren2016gprmax}; 3) we train and fine-tune the Faster-RCNN based on the pre-trained weights and test on both simulated and real data in order to prove the effectiveness of the proposed approach.

The remainder of this paper is organized as follows. Section \ref{sec:data} first presents the real collected GPR data as well as the simulated radargrams used in this work. In Section \ref{sec:method}, we describe the proposed approach to adapt the Faster-RCNN for hyperbola detection from our data. We then provide some preliminary detection results in Section \ref{sec:result}. Finally, Section \ref{sec:conclusion} concludes the paper and discuss some on-going work as well as future perspectives.

\section{Data sets}
\label{sec:data}
\subsection{Real collected data}
\begin{figure}[!ht]
\centerline{\begin{minipage}[b]{0.98\linewidth}
  \centering
  \includegraphics[height=38mm]{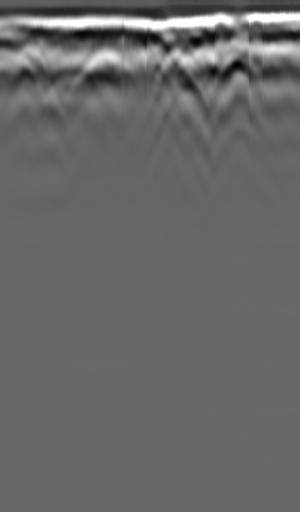}
  \includegraphics[height=38mm]{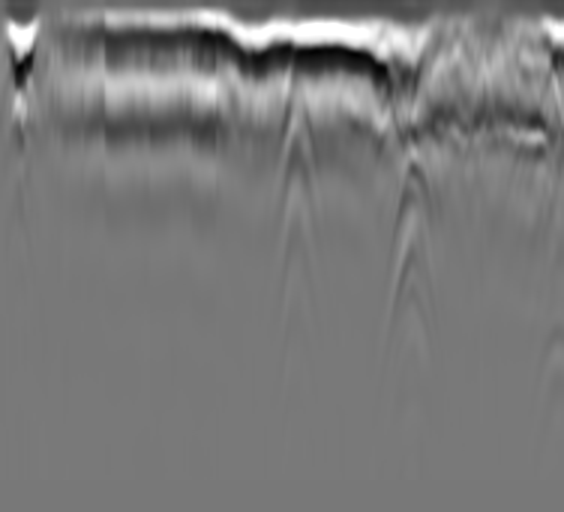}       
\end{minipage}}
\caption{Examples of real GPR radargrams.}
\label{fig:gpr_real}
\end{figure}

In this work, we exploit about 100 real B-scan data recently collected from several sites in France using a GPR antenna of 300MHz. Each acquisition has a time range of 100ns in depth and can penetrate up to 7 meters. Within these data, hyperbola signatures were recorded from the reflection of electromagnetic signal on buried objects with different shapes and materials. Only some are clear and well-shaped, while most of them are weakly contrasted, asymmetric and perturbed by noises and clutters caused by the soil heterogeneity and the change of impedance between different subsurface layers. Figure \ref{fig:gpr_real} shows two examples of real data from which we can observe some hyperbolas of various sizes that suffer a lot of noises and clutters.

\subsection{Simulated data}
\begin{figure}[!ht]
\centerline{\begin{minipage}[b]{0.98\linewidth}
  \centering
  \includegraphics[height=38mm]{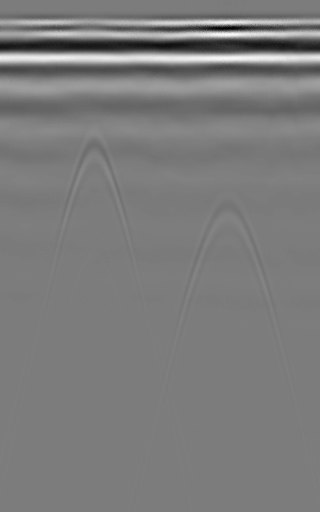}
  \includegraphics[height=38mm]{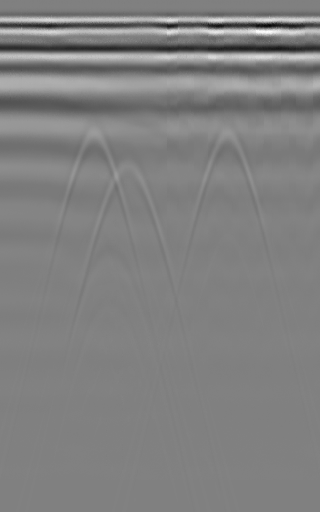}       
\end{minipage}}
\caption{Examples of simulated GPR B-scans using gprMax.}
\label{fig:gpr_sim}
\end{figure}

Due to the lack of real data for training, more GPR radargrams were generated to simulate different scenarios using the gprMax toolbox \cite{warren2016gprmax}. Various configurations were considered where objects of different sizes and materials were placed at different positions and depths. The same antenna frequency of 300MHz and same time range of 100ns were set for simulation. The simulated images were then added some noises which were estimated from the real data. Figure \ref{fig:gpr_sim} shows two examples of simulated radargrams where hyperbolas (generally well-shaped) are intersected and have crossing prongs. In this work, 50 simulated radargrams were simulated and added to the previous real data to perform and evaluate the proposed framework.

\section{Application of Faster-RCNN}
\label{sec:method}
\begin{figure*}[!ht]
\centerline{\begin{minipage}[b]{0.99\linewidth}
  \centering
  \includegraphics[width=0.65\linewidth]{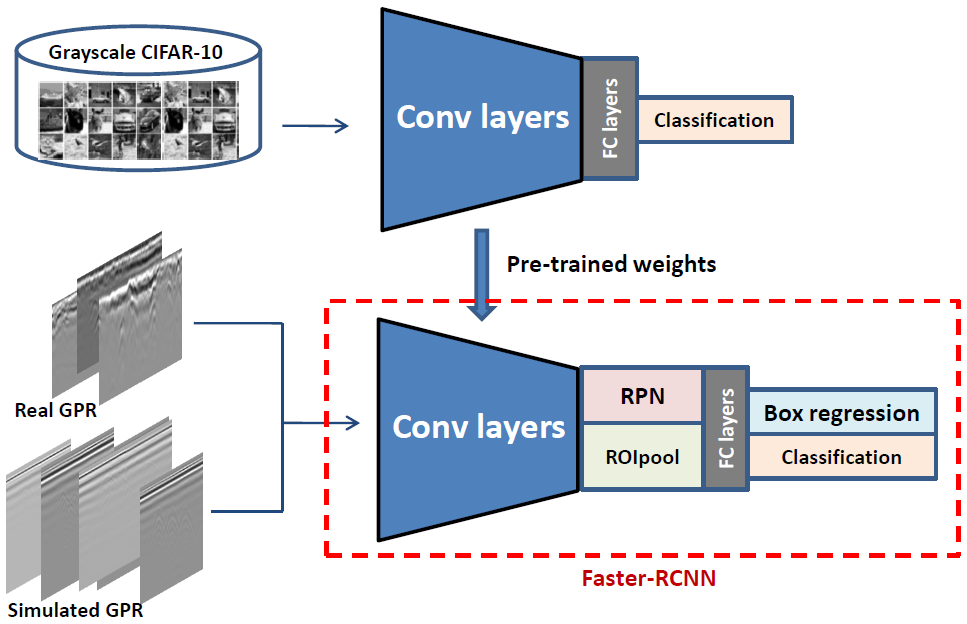}
\end{minipage}}
\caption{Proposed framework for buried object detection from GPR B-scan images.}
\label{fig:framework}
\end{figure*}

The proposed approach consists of two main stages which can be observed from Figure \ref{fig:framework}: 1) pre-train a designed CNN on the grayscale Cifar-10 database;  2) train and fine-tune the Faster-RCNN (based on pre-trained CNN weights) using both real and simulated GPR data. We now describe each of them in details.
\subsection{Pre-training a CNN on the Cifar-10 database}
CNNs are generally comprised of convolution layers followed by pooling layers and fully-connected layers. Our defined CNN simply includes 3 convolution layers of 16, 32 and 64 filters of size $5\times 5$ pixels (each one is followed by a ReLu activation and a max-pooling layer of size $2\times 2$ pixels) and one fully-connected layer of 64 neurons. As recommended in \cite{reichman2017some}, the Cifar-10 was chosen (instead of ImageNet) since the image size is small ($32\times 32$ pixels),  which approximates the size of hyperbolas within the studied GPR images. It can be trained faster and easier on a personal computer with limited GPU. Also, in order to match single-channel GPR data, it is better to train the CNN on grayscale Cifar-10 instead of the color database.
\subsection{Training the Faster-RCNN on both real and simulated GPR data}
The Faster-RCNN involves two main components: the Region proposal network (RPN) and the Fast-RCNN \cite{ren2015faster}. For a brief description, the role of RPN is to generate a set of region proposals while the Fast-RCNN (including a classifier and a box regression operator) detects whether each region is an object or not. They both share the same weights from the previously pre-trained CNN. For more details about the Faster-RCNN framework, readers are referred to \cite{ren2015faster}. 

As shown in Figure \ref{fig:framework}, both real and simulated GPR images were exploited to train the Faster-RCNN using the pre-trained CNN weights. All implementations in this work were carried out based on the MATLAB Neural Network toolbox using a PC with a GPU compute capability 5.0.
\section{Preliminary results}
\label{sec:result}
In order to evaluate the proposed approach on both simulated and real GPR data, three scenarios were tested:
\begin{enumerate}
\setlength{\itemsep}{0pt}
\setlength{\parskip}{0pt}
\setlength{\parsep}{0pt}
\item Train and test on simulated data;
\item Train and test on real data;
\item Train on simulated + real data, test on real data.
\end{enumerate}
We note that training and test samples were split so that they were well separated. For real data, we used 60 radargrams for training and 40 for testing. For simulated data, 40 were set for training and 10 for testing. We now provide some preliminary results with qualitative assessment on each test scenario in order to confirm the effectiveness of the proposed framework. 
\subsection{Performance on simulated data (scenario 1)}
\begin{figure}[!ht]
\centerline{\begin{minipage}[b]{0.98\linewidth}
  \centering
  \includegraphics[height=40mm]{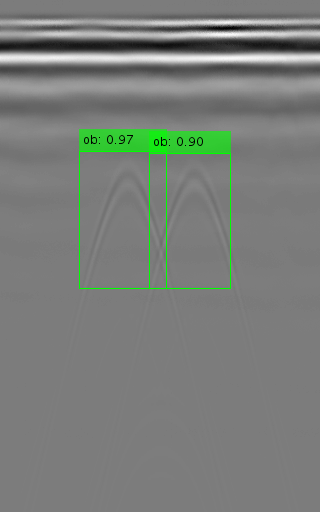}
  \includegraphics[height=40mm]{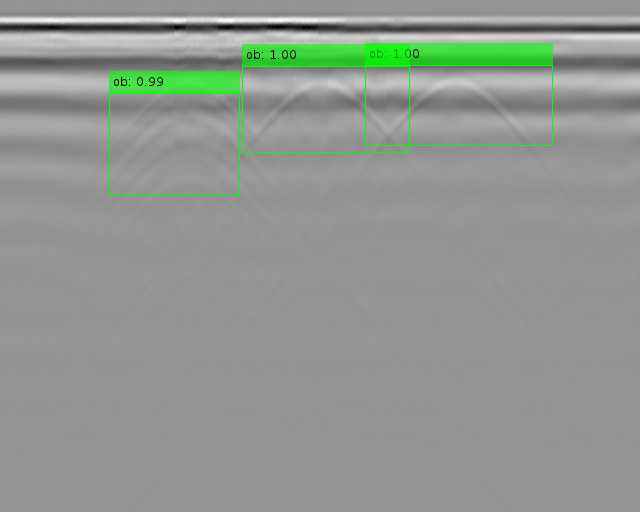}       
\end{minipage}}
\caption{Detection results on simulated data.}
\label{fig:res_sim}
\end{figure}

The first scenario aims at quickly evaluating the proposed approach only on simulated data. Some detection results are reported in Figure \ref{fig:res_sim} where hyperbolas were detected with high confidence scores. In general, our experiments showed that the framework can cope well with this scenario in order to provide good performance on simulated images.
\subsection{Performance on real data (scenarios 2 and 3)}
The other two scenarios were experimented when working on real data. The motivation is to prove that by adding more simulated radargrams for training, the detection framework could provide better performance compared to the case where only real data (with limited quantity) were exploited. In addition, one classical recognition technique called cascade object detector (COD) based on the Viola-Jones algorithm \cite{viola2001rapid} (which was exploited in \cite{maas2013using}) was implemented for comparison. 

\begin{figure*}[!ht]
\begin{minipage}[b]{0.49\linewidth}
\centering
  \includegraphics[height=44mm]{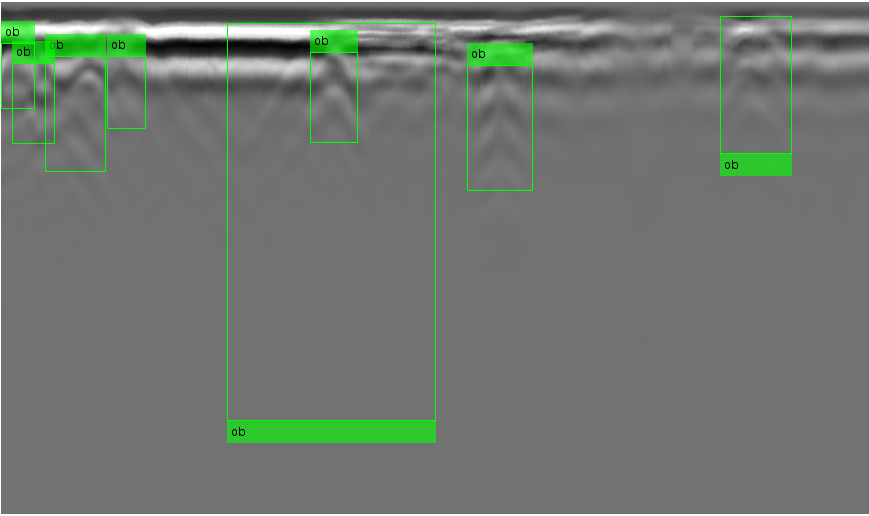} \\
  \footnotesize{(a) COD - HOG}
\end{minipage}
\hfill
\begin{minipage}[b]{0.49\linewidth}
\centering
  \includegraphics[height=44mm]{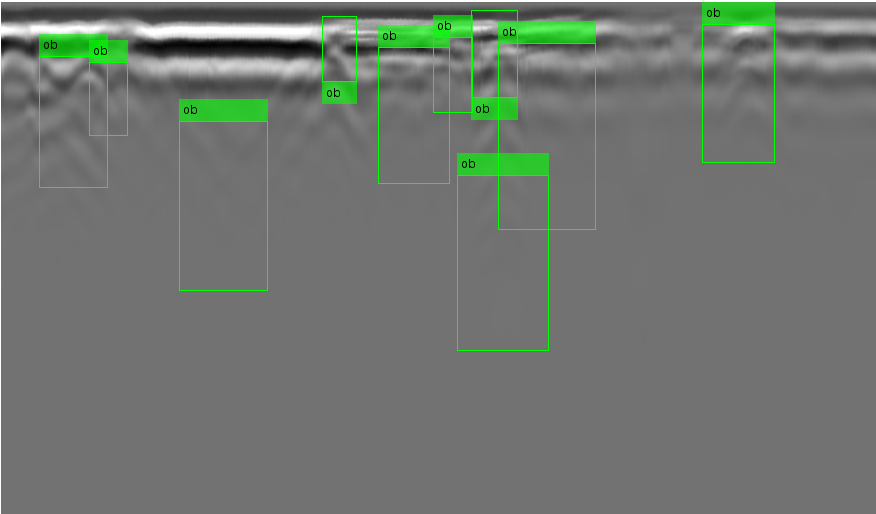} \\
  \footnotesize{(b) COD - Haar-like}
\end{minipage}
\vfill
\vspace{1mm}
\begin{minipage}[b]{0.49\linewidth}
\centering
  \includegraphics[height=44mm]{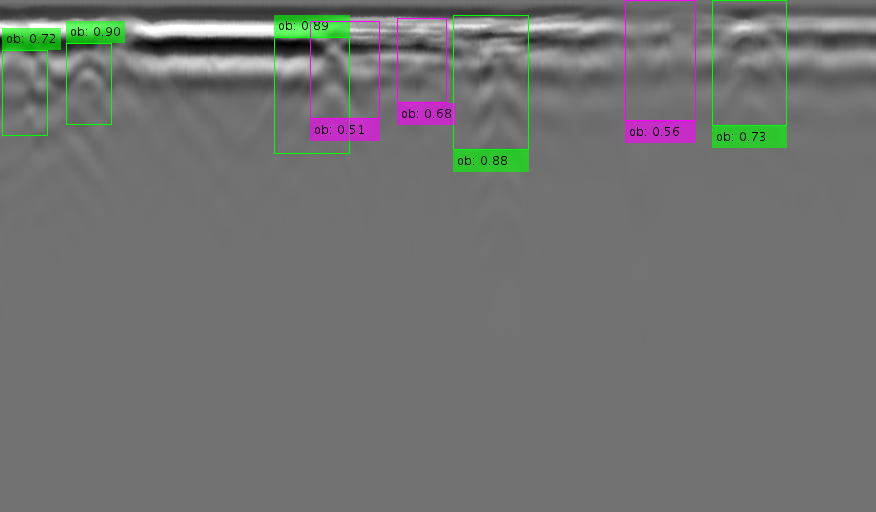} \\
  \footnotesize{(c) Faster-RCNN scenario 2}
\end{minipage}
\hfill
\begin{minipage}[b]{0.49\linewidth}
\centering
  \includegraphics[height=44mm]{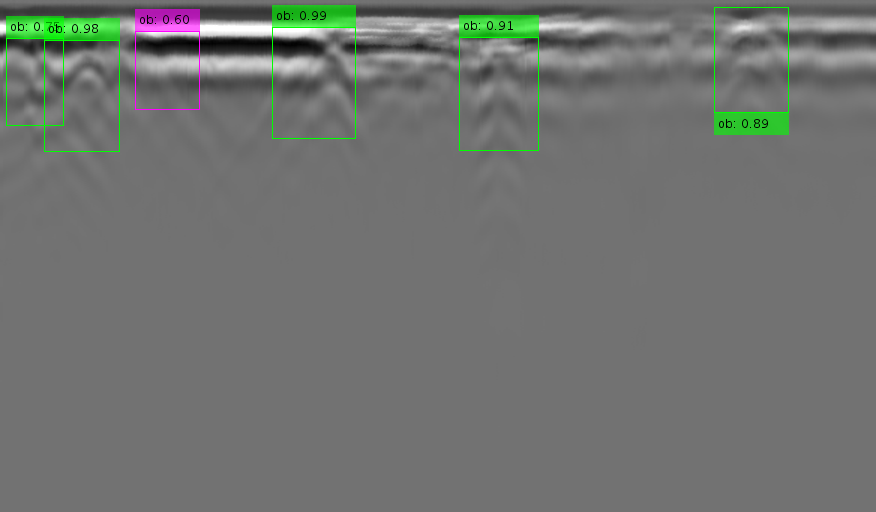} \\
  \footnotesize{(d) Faster-RCNN scenario 3 (best)}
\end{minipage}
\caption{Detection results on real GPR data yielded by the Faster-RCNN (c and d) compared with the COD (a and b). For Faster-RCNN, detected boxes with high confidence index ( $\geq$ 0.7) are marked in green; those with index  $<$ 0.7 are in magenta.}
\label{fig:res_real}
\end{figure*}

In Figure \ref{fig:res_real}, we compare the detection results on a real GPR image yielded by our two scenarios compared with the COD based on HOG and Haar-like features. Here, results are shown without any post-processing technique. As we can observe, the Faster-RCNN can provide better performance compared to COD method which yielded unstable object bounding boxes and more false alarms. Importantly, adding more simulated data could provide detection results with higher accuracy and confidence (more good detections and fewer false alarms), which validates our intention and confirms the effectiveness of the proposed scheme.
\section{Conclusion and further work}
\label{sec:conclusion}
We have applied the well-known Faster-RCNN framework to the detection of buried objects from GPR B-scan data. By combining both simulated and real collected radargrams for training, the proposed technique can provide good performance on tested real data and considerably outperforms detectors using classical features such as HOG and Haar-like. Therefore, it becomes promising to deal with GPR data with limited training samples.

Although the effectiveness of the proposed scheme has been qualitatively observed, our on-going work now focuses on quantitative evaluation for a better validation. Moreover, further work related to the detection of hyperbola coordinates (apex and prong) is necessary for a fine and accurate localization of detected objects.
\begin{small}
\bibliographystyle{ieeetr}
\bibliography{RefAbrv,RefGPR}
\end{small}

\end{document}